# Reconstructing vehicles from orthographic drawings using deep neural networks

Robin Klippert
Coburg University Of Applied Sciences
Germany
Robin.Klippert@stud.hs-coburg.de

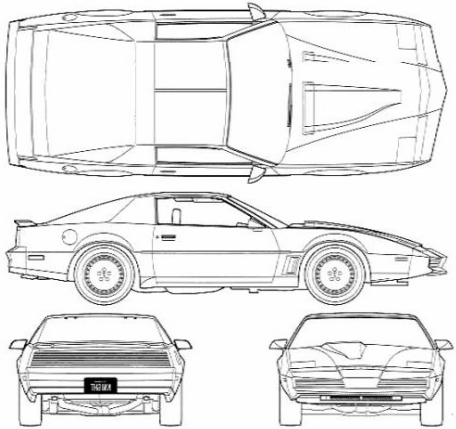
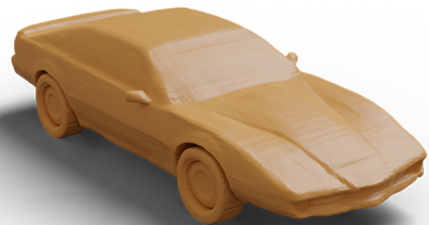

## ABSTRACT

This paper explores the current state-of-the-art of object reconstruction from multiple orthographic drawings using deep neural networks. It proposes two algorithms to extract multiple views from a single image. The paper proposes a system based on pixel-aligned implicit functions (PIFu) and develops an advanced sampling strategy to generate signed distance samples. It also compares this approach to depth map regression from multiple views. Additionally, the paper uses a novel dataset for vehicle reconstruction from the racing game Assetto Corsa, which features higher quality models than the commonly used ShapeNET dataset. The trained neural network generalizes well to real-world inputs and creates plausible and detailed reconstructions.

## 1 INTRODUCTION

While it is often a simple task to find pictures of a real-world object on the internet, finding a 3D representation of an arbitrary object often turns out to be rather expensive. This is because manual 3D modelling of complex objects is time-intensive since it is difficult to algorithmically solve this task from pictures. Approaches like photogrammetry require detailed photographs from all viewing angles in controlled lighting conditions, which are only possible to acquire if the object is accessible in the real world.

However, it is often possible to find orthographic drawings (also called blueprints) of mechanical objects, such as cars, trains, planes etc. on the internet. These drawings capture important features and proportions from multiple directions which usually makes them a reference point for manual modelling. Since a human can model a mesh from this representation, the question arises if this kind of input data can also be used to infer the mesh automatically.

This paper therefore explores the current state-of-the-art deep learning algorithms for vehicle reconstruction from multiple orthographic drawings since recent advances in the area of deep learning show promising results.

## 2 MOTIVATION

As previously mentioned, the creation of complex 3D objects is currently mostly done by humans, which makes the process rather expensive and time-intensive. Algorithmically generating a final- or intermediate result could therefore lead to large time savings. Even if the prediction from the neural network cannot be used directly, it can still be used as a base or starting point to speed up the manual process.

Another interesting field could be rapid prototyping: It would be possible to quickly draw a sketch depicting the object by hand, which can then be converted into a 3D printable model. This paper proves that this is possible by 3D printing a vehicle, which is generated from a blueprint by a neural network. This task can be solved well by implicit function approaches, which predict watertight meshes by design.

Object reconstruction from images can also be useful in autonomous robots or cars, where it is often necessary to sense the surrounding area from cameras. This is crucial for collision detection, which is a major focus of autonomous vehicle development. Robust and fast 3D reconstruction is therefore useful in this area.

## 3 PREVIOUS WORK AND SURFACE REPRESENTATIONS

An ideal technique to solve the reconstruction problem of this paper should have the following properties, ordered by importance:

- Multi-view image input
- High-resolution reconstruction
- Usage of local details
- Low VRAM consumption
- Fast inference

Multi-view image input and high-resolution reconstruction should be obvious requirements. It should therefore also be possible to leverage local details in the images. Many methods compress the image into a compact feature vector, which can intuitively be seen as a global descriptor of the image. Local details cannot be contained in this description. Low VRAM consumption makes the network accessible to researchers with only consumer cards available. Fast inference has the lowest priority on the list because the time saved with automatic inference is still large compared to manual modelling.

Volumetric techniques divide space into a 3D regular grid of $size^3$. These methods are usually limited by their memory footprint and dont allow high resolution reconsturctions. The general idea of voxels makes little sense in the reconstruction of cars because only the detail of the surface is important. One interesting paper using voxels is SurfaceNet [9], which uses voxel slices of a bigger mesh to create a high-resolution object. They project image slices from multiple images into a small voxel grid, to take multiple views into account. The reconstruction allows high detail compared to other voxel methods, but the result has holes in it.

Parametric representations map a 2D domain into 3D space: "Spherical parametrizations and geometry images are the most commonly used parameterizations. They are, however, suitable only for genus-0 and disk-like surfaces. Surfaces of arbitrary topology need to be cut into disk-like patches and then unfolded into a regular 2D domain. Finding the optimal cut for a given surface, and more importantly, findings cuts that are consistent across shapes within the same category is challenging." [6][p.8]. Papers that use this approach often fail to produce smooth surfaces [31][5][25]. SurfNet [25] uses spherical projection images and voxelizes the meshes into a $128^3$ grid as a pre-processing step, which limits the quality.

Another option is to use template deformation. A general template is deformed to match the desired shape. This approach is more fitted for e.g. human reconstruction, which are very similar from a general point of view. ARCH [7] uses a deformation approach to deform an arbitrary pose of a human to a canonical T-pose for further rigging and animation. The final result is generated as an implicit surface.

Point based methods are also a common choice for 3D reconstruction. An object is defined by a set of points $S = \{x_i, y_i, z_i\}_{i=1}^{N}$ of $N$ points. This representation also does not easily fit into common convolutional neural networks, since it is not a regular structure. The creation of depth maps can also be put in the intermediate representation category.

MVPNet [30] for example fails to generate any detail. It also only takes one view into account and cannot leverage local detail. Point-based methods also have the disadvantage of requiring an additional fusion step to create a mesh, this is often solved with screened Poisson surface reconstruction [11].

Intermediate representations do not directly represent 3D objects, but usually create two-dimensional outputs, like depth- or normal maps, which are then used in a post-processing step to create the final model. These methods are convenient since deep learning techniques are well researched in the two-dimensional domain, especially image-to-image translation.

Lun et al. [17] is essentially trying to solve a problem identical to ours: They reconstruct a character from multi-view sketches. Soltani et al. [28] trains an autoencoder that generates 20 depth maps from one or multiple input depth maps. The results capture the general shape well but fail to add any detail. This paper can also not incorporate any local detail by design.

Umetani and Nobuyuki [29] also train auto-encoder to generate depth maps of cars, while it is focused on generating novel car shapes from latent space. The depth maps are converted into a mesh by a shrink-wrapping approach, which generates nice details like the front grill. This method cannot leverage local detail as well, as the decoder only has access to the latent space. It also requires manual pre-processing of over 600 cars, so antennas, mirrors, etc. could be removed to make the shrink wrapping work. Smith et al. [27] uses both a volumetric and intermediate approach. They take a single image as an input and construct a low-resolution voxel representation. The authors then render multiple depth maps of this object which are processed by an ESRGAN [32] super-resolution network, which upscales each depth image. The images are then used to refine the original voxel representation. The result is not very detailed, however. Kar et al. [10] projects multiple input images into a voxel grid, from which either a voxel or depth map representation can be extracted. This approach is similar to SurfaceNet [9]. Their voxel results lack detail, while the depth map results have detail, but the overall shape is bumpy.

Bi et al. [3] uses six RGB input images to predict depth-, normal-, specular albedo-, roughness-, and diffuse albedo maps. Since it also creates diffuse and specular textures the results can be rendered immediately, which share nice reconstruction detail. It is questionable however if this technique also works when the inputs are four black-and-white images, which do not contain as much information as an RGB image. The depth map projection approach also suffers from a lack of expressiveness, since only surfaces that are visible from a certain viewpoint can be modelled.

Another technique for surface reconstruction is implicit functions, which can currently be considered to be state of the art. Most papers of 2019 and 2020 that generate high quality meshes are using implicit functions [22][23][7].

The general idea is to train a function $F$ that takes a point $p \in [0, 1]^3$ and calculates the signed distance to the surface [4]. This creates a regression problem, but some methods also simplify it to a classification problem [22].

This technique saves memory compared to volumetric approaches since it does not require the entire geometry to be loaded, instead,

it is enough to only load signed distance samples of it. During inference, the space is sampled in a regular grid to create a field of signed distance values. The mesh can be extracted from the volume with algorithms such as Marching Cubes [16] or Dual Contouring [24].

Chen et al. [4] proposes IM-NET, which generates an implicit field from a single image or a latent vector. The result has nice visual properties, the surface is smooth and has a average amount of details. The disadvantage of their approach is the preprocessing step into voxels of size $128^3$, which removes some details in the model. Due to the single image encoder, they can only process a single image which is compressed into a latent vector. So the decoder cannot access local features inside the image. They also propose a generative adversarial network that generates novel shapes in latent space. [18] proposes occupancy networks, which is similar to the paper above. This means it also shares the mentioned weaknesses of the previously discussed approach.

Liu et al. [14] develops an approach to construct implicit surfaces from an image without 3D supervision. This differentiable rendering approach only uses very small resolution images and creates overly smooth and low detail shapes. Atzmon et al. [1] proposes a technique to generate an implicit field with a neural network on a raw triangle soup.

DeepSDF [20] also generates an implicit field as its output, however, they design their network for novel shape generation and shape completion, so they do not use an image reference as input. This allows them to propose the idea of an auto-decoder. They do not use an encoder, instead, they assign a random latent vector to each training sample and optimize the latent vector with backpropagation to fit the samples of the signed distance field.

PIFu [22] and PIFuHD [23] are the most valuable papers in this list because they solve a problem with many previous approaches. ARCH [7] also seems to be directly inspired by this work. While previous approaches produced nice looking shapes with average detail, they always utilized a global latent vector encoding, which compresses the image into a compact representation. This makes it impossible to access fine, local details in the image. PIFu solves this problem by utilizing a fully convolutional image encoder, that converts the original image into a feature map. An hourglass network [19] is used as an image encoder. PIFu transforms each sample from 3D space into 2D image space before the predictor is queried. This requires that for each image the camera parameters have to be known, but it also allows the system to access local features in the feature maps by sampling features from the grid. The updated HD variant [23] allows higher resolution inputs with additional changes.

PIFu satisfies the most important requirements: It enables multi-view image inputs, allows high-resolution reconstruction and leverages local details. The GPU memory consumption is also no problem for consumer GPUs.

## 4 VIEW EXTRACTION AND IDENTIFICATION

This section deals with the problem of view extraction and identification. Two algorithms are implemented to solve this issue: A line cutting algorithm and a contour-based approach. The line cutting algorithm tries to find rows or columns of pixels that are empty. It then repeatedly cuts the image into smaller sections until the expected number of views is found. The contour-based algorithm uses OpenCV's external contour functionality to find each view. The line cutting algorithm is tried first, if the number of found images is not four, the alternative contour algorithm is used. The resulting views are then partially classified by their size and the result is displayed on the web interface for the user to check and finalize. If both cutting algorithms fail, the user has to manually define the bounding boxes inside the image on the web interface.

The view identification problem is only solved partially in this paper. The implementation sorts the collection of images by their size and filters out the four largest ones. The remaining images are then sorted by their width, which should return the side and top view as the first two images. The image with the larger height is picked as the top view, which is the case for most vehicles. The remaining two images, therefore, have to be either the front or back view, which is left for the user to classify, which is also the case for the viewing direction of the side and top view.

It would also be possible to solve this task with a neural network that predicts bounding boxes. However, the manual algorithms work reliably enough in practice. Since this step is crucial to reconstruct the vehicle correctly, it would probably still be the best choice to supervise the result before it is processed further, even if a neural network is used. A poorly cut blueprint or a misclassification creates an incorrect result.

## 5 NETWORK ARCHITECTURE

As previously mentioned, the proposed system is based on PIFu [22]. Numerous changes have been made to the architecture to improve the results for multi-view vehicle reconstruction. Important changes are listed in Tab. 1.

The coordinate system is changed from a real-world scale, which reconstructs humans in their actual height, to a normalized system. The vehicles are scaled to a length of 1 and placed at the origin, since reconstructing the actual real-world size is of little interest.

An important change is the switch of average feature pooling to feature concatenation. The original implementation calculates each view in parallel until a specific layer inside the classifier is reached. Then the latent vectors from all views are averaged and passed through a final layer, which makes the final decisions. This has the advantage of being flexible with the number of input views, so the architecture does not change between a single or multi-view input.

However, the average operation also causes a loss of information. This can be best explained by looking at the front and back view. These views are independent of each other in the real world, the front of a car does not influence the back of it. But in the original implementation, features from the front and back view are averaged for each point, which does not make sense in our case. The classifier also does not have access to information from multiple views in an early stage, so it can only work with information from a single view up until the average operation.

A theory for why this is not a problem for human reconstruction is that the front and backside of a human are closely related, while a vehicle is a far bigger object in comparison, with little similarity between the front and backside.

| Design choice | Original | Adapted |
|---|---|---|
| Coordinate system | Real-world units | Normalized space |
| Multi-view aggregation | Average pooling | Concatenation |
| Number of input views | Adaptive | Fixed |
| Information fed to MLP | Feature stack, depth per view | Feature stacks, world space coord. |
| Multi-view image size | Fixed | Dynamic |
| Sampling timing | During training | Before training |
| Sampling strategy | Uniform | Adaptive |
| Number of sample points | 5000 | 20000 - 25000 |
| Sampling/reconstruction space | Fixed, entire space | Individual bounding box |
| Losses | Occupancy | Truncated SDF, normals, edges |
| Batch size | > 1 | 1 |

Table 1: Design choices different to the original PIFu implementation

The system is therefore changed to concatenate image features from all views into a combined feature vector for each point, which is then given to the classifier together with the world space coordinate. Using the world space coordinate is the simplest choice since all models are normalized. This change allows the classifier to make decisions early on based on information from multiple views without any loss of information. The disadvantage is that the architecture is now fixed to four views, but this is not a problem in our case.

Another important change is the switch from fixed image size to a dynamic one, since only a portion of the image is used by the vehicle. This is especially true for the front and back view. Using the original square images wastes important memory and computation power. The code is therefore changed to iteratively calculate features from each image after it has been cut according to its bounding box. Even though we lose parallelism by this approach, the training time is nearly cut in half and a large amount of memory is saved. In experiments, the memory consumption fell from around 7,5GB to around 4,5GB. As a result of this, we can use more samples during training and a larger network with more parameters, which in turn improves the final result. The mentioned adaptive sampling strategy in Tab. 1 receives its own section.

The final change is the switch from classification (occupancy) to regression. This causes a smoother reconstruction result because classification can lead to sharp changes in predicted signed distance values, which in turn causes a staircase effect in the Marching Cubes [16] reconstruction result. This problem could not be fully resolved and is still present in some reconstructions. Fig. 1 shows the final architecture: Note that each image is encoded iteratively, which allows the use of different image sizes.

Since cars are highly reflective, their first- and second order derivates play a vital role in visual appearance. Thus, we also condition the network to predict correct normals and edges. PIFuHD [23] also uses normal loss, but does not utilize edge loss (second-order derivative). The batch size is also fixed to one since experiments show that the reconstructed vehicles contain sharper details compared to vehicles trained with larger batch sizes. Using a batch size of one also allows to utilize a larger network combined with more training samples.

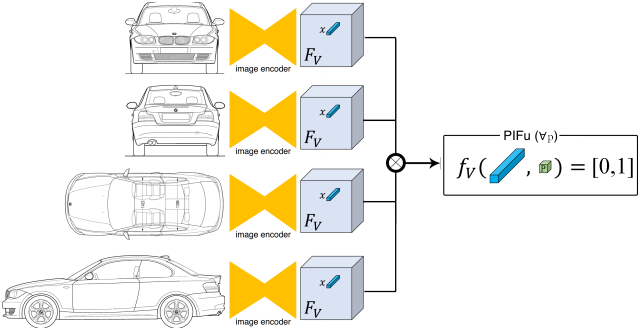

Figure 1: Overview of updated architecture

Changes have been made to the Hourglass network, see Tab. 2. The amount of stacks has been reduced to two, since doing a lot of computation in image space in a multi-view setup is of little use. Only one downsampling step is applied, which cuts the initial resolution in half instead of a quarter. This allows a more detailed reconstruction with higher resolution. The number of downsampling operations inside each module has been increased to five, which gives each queried sample a broader field of view.

Two networks are trained on 512 and 256 pixel images. This refers to the maximum length of the top or side view since we are not using quadratic images. The network uses a feature depth of 256 for the smaller 256 pixel network and a shallower 128 network for the larger 512 pixel network for memory reasons. The 256 pixel network can be used on blueprints with higher resolution to create a more detailed reconstruction.

The authors of the hourglass network [19] also propose an intermediate loss function approach. This is also implemented in the original code but caused less detailed results. This is probably because the classifier gets trained with features it does not see in the evaluation phase. This intermediate loss scheme is therefore removed.

| Hourglass network settings | Original | Adapted |
|---|---|---|
| Intermediate Hourglass loss | Yes | No |
| Number of stacks | 4 | 2 |
| Initial downsampling steps | 2 | 1 |
| Internal downsampling steps per stack | 2 | 5 |
| Input resolution | 512x512 | Largest dim. 512 or 256 |
| Feature resolution | 128x128 | Largest dim. 256 or 128 |
| Feature depth | 256 | 128 or 256 |

Table 2: List of changes inside the hourglass image encoder

## 6 DATASET PREPROCESSING AND ADVANCED SAMPLING STRATEGY

While many methods rely on ShapeNET, using vehicles from the Racing Game Assetto Corsa provides higher quality models. These models are directly loaded into Unity for rendering and further processing.

Reconstructing the wheels together with the vehicle does not yield good results, since rims are quite detailed with thin structures. They are also relatively small compared to the entire car, so the image area covering them cannot capture this detail very well. The wheels, therefore, get replaced with a default wheel, which is scaled to fit the original. It is important to note that the generated blueprints contain the original wheels, the default wheel is only used when the signed distance function is calculated.

To create a complete blueprint a set of four views are needed, each rendered as a drawing. The views depict the front, back, side and top view respectively. This is achieved with an edge detection filter inside Unity as a post-processing effect. This highlights sharp edges but does not capture lines where different parts meet. Therefore, we also extract an image colouring each part differently, which also receives an edge detection filter. Extracting each part is non-trivial since many models combine different parts into a single mesh. The line and part images get multiplied together to create improved drawings.

While these blueprints get close to a real-world blueprint in a macro view, they are too perfect on a pixel level. Actual blueprints may contain noise, compression artefacts or other lines that are not part of the geometry. So before the network receives the input, each blueprint gets augmented by adding random noise, compression artefacts and lines. The same blueprint can be augmented in multiple ways to show different possible blueprints of the same car, forcing the network to also deal with these noisy inputs. This step is crucial to achieving good generalization properties. To further enhance the augmentation, we also save the views of each car with its windows removed, showing the interior.

The final preprocessing step is to create a signed distance field with normal and edge data from the mesh to generate an implicit surface. This step is rather complicated since the dataset does not contain watertight models.

The authors of PIFu[22] emphasize the importance of sampling strategy in their paper [22][p. 12]. Their experiments show that uniformly sampling the entire space leads to poor results since the majority of samples are in empty space. This causes the decision boundary to be less sharp and loses local details. They, therefore, combine uniform samples with surface samples, which are shifted by a standard deviation $\sigma$.

However, the surface samples are evenly distributed over the entire model. This is a valid choice for human reconstruction since details are evenly spread on clothed humans, with no large flat areas. This does not hold true for vehicles, which have areas with many details (e.g. the front grill) mixed with large flat areas of little detail (e.g. the side doors or roof). This causes oversampling in less detailed areas and undersampling in highly detailed areas. A more adaptive sampling strategy is therefore developed to improve the distribution of samples.

As previously mentioned the python package mesh_to_sdf [12] is used as a starting point. This package implements the system which is proposed in DeepSDF[20]. Multiple virtual cameras are placed around the model to render a depth and normal image, which essentially creates a scan of the model.

An example of the generated data can be seen in Fig. 2. The left part depicts the signed distance field where red points are considered to be outside and green points are considered to be inside. The centre image shows the generated normal samples and the right image shows the generated edge samples. A brighter colour shows a sharper edge.

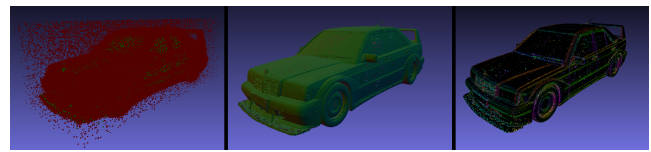

Figure 2: Example of generated ground truth SDF field, normal samples and edge samples

Note that the viewing angle makes the generated data look uniformly distributed, while in reality, it is not. The following paragraphs explain the different parameters that play a role in the distribution function.

A good distribution of surface points should place

- ... only few samples on the undercarriage
- ... more samples near sharp edges
- ... fewer samples on flat areas
- ... more samples in areas that deviate from the outer hull
- ... more samples on thin structures

The formula for calculating the weight for each point is therefore defined as

$$weight = wEdge \cdot wNormal + wHullDist \cdot wHullDistNormal + wThickness. \quad (1)$$

The weight for each point is then normalized such that the sum over all points accumulates to 1, which creates a probability distribution. The different factors are additionally balanced by multiplication and clipping values, which have been omitted for brevity. The effects of each factor can be seen in Fig. 3.

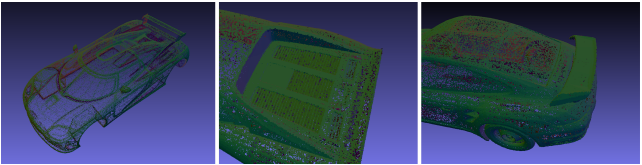

**Figure 3: Left: Edge weighted sampling, Centre: Hull distance weighted sampling, Right: Thin structure weighted sampling**

The factor wEdge is larger when the sample is near a sharp edge and lower on a flat area. We obtain this value by applying a Sobel filter on the rendered normal image. This factor already mostly satisfies the first three requirements for a good distribution of samples. Samples on the undercarriage are removed if this area is perfectly flat, but there are still too many samples if the model contains details like the exhaust and suspension. We, therefore, multiply wEdge with wNormal, which is defined as

$$wNormal = \begin{cases} 1, & \text{if } relH < 0.05 | relH > 0.5 | normalY > -0.95 \\ relH, & \text{otherwise.} \end{cases} \quad (2)$$

relH is 0 at the bottom of the tires and 1 near the roof. By defining wNormal like this, we map down facing samples that are positioned between 0.05 and half the height of the car to their relative height. We do not start at 0 so we do not remove samples on the bottom of the tires.

The next requirement for a good sample distribution is to sample more heavily in occluded areas, which have a large distance to the outer hull. The trained network tends to mainly follow the outer hull of the vehicle, which works well for many cars but fails for some vehicles.

The general idea is to calculate the distance from each point to the outer hull and increase the weight for points with large distances. We can intersect the silhouettes of all four views to create a rough approximation of the vehicle. We can then convert the generated voxels into a point cloud and use a k-d tree to calculate the distance for each sample to the outer hull.

This results in the wHullDist factor, which gets multiplied by wHullNormalDist, which is a heavily scaled-down version of wNormal. wHullNormalDist is used to avoid adding too much weight for samples that are far away from the outer hull but point downwards.

The final factor is wThickness, which is large for thin structures such as wings. Experiments show that thin structures can be challenging because they require quick changes in signed distance values to be reconstructed correctly.

The general idea is to calculate the distance between different point clouds, which have been created by splitting the entire normal point cloud by the sign of each point for each axis. We again use a k-d tree to find the closest point for each axis between the two point clouds. We then set the weight for each point to the inverse of the returned distance if the angle between them is larger than a specified threshold. So points with close distances receive a large weight if they point in largely different directions.

## 7 RESULTS AND LIMITATIONS

This section presents the final results of this thesis for the pixel-aligned implicit function approach. Fig. 4 shows reconstruction results for the 512 input size network.

The images are resized to the ideal resolution if they are 20 percent larger or smaller than the size the network is trained on. The largest volume is extracted, which gets rid of small floating artefacts.

It is visible from the results that the network managed to generalize across different vehicle shapes and different blueprint types. The 512 pixel network can generate reconstructions with higher detail, which can be seen in Fig. 5.

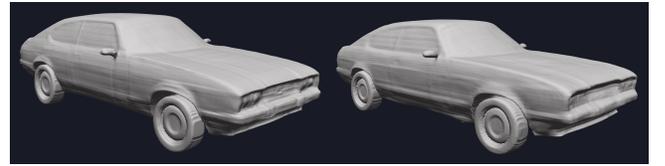

**Figure 5: Difference in detail between the 256 and 512 input size network**

While both networks do a good job of filtering out measurements and interior details in the front, back and top view, the side view is prone to visible seats. The 512 input size network puts a strong imprint into the side, which also happens for the 256 input size network, but less pronounced. Fig. 6 show an example of this. The augmentation pipeline has to be improved to avoid this misinterpretation.

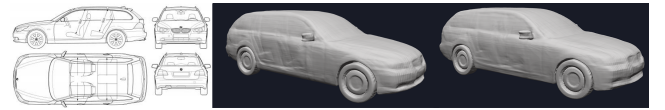

**Figure 6: Reconstruction artefacts from lacking augmentation, Left: Input blueprint, center: 512 input size network, right: 256 input size network.**

Another challenging area is thin structures, such as rear wings or the windshield for cabriolets. While the sampling strategy already accounts for this, there is no guarantee that these structures are always attached to the vehicle. Since floating objects are removed, rear wings can sometimes be missing. Fig. 7 shows an example of this problem. The right reconstruction result is from the 512 input size network, while the surface is extracted at the normal 0.5 threshold. The centre image shows the result for the 256 input

Reconstructing vehicles from orthographic drawings using deep neural networks size network using the same threshold. It can be seen that the smaller resolution network reconstructs the wing partially, but not completely. It is also possible to reduce the Marching Cubes [16] extraction threshold slightly to 0.45, which reconnects the wing but also slightly thickens the entire model.

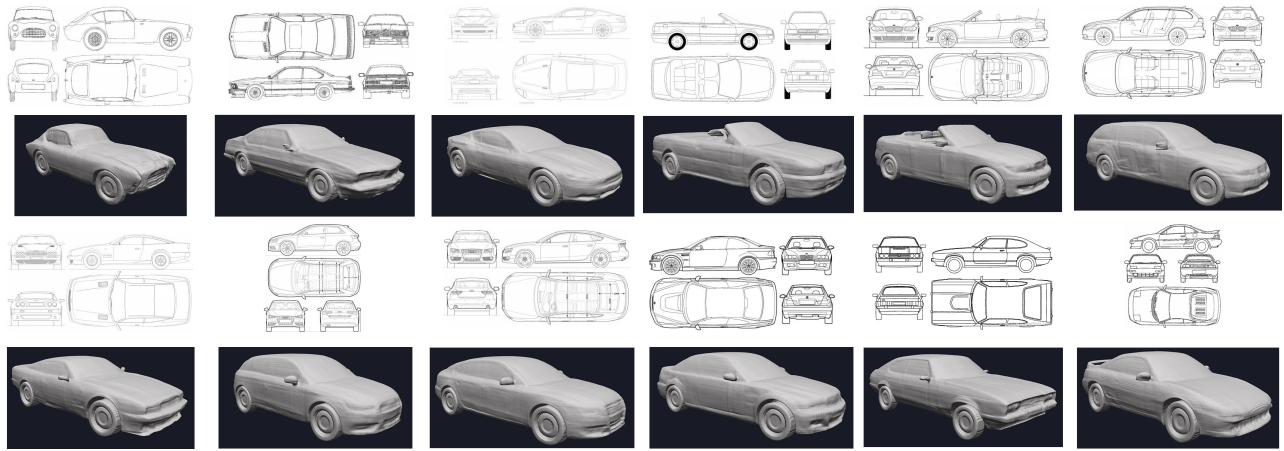

**Figure 4: Reconstruction results using the 512 input size network**

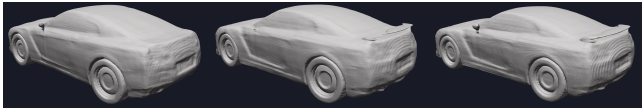

**Figure 7: Reconstruction failure for rear wing, Left: 512 input size network, Center: 256 input size network, Right: 256 input size network with 0.45 threshold**

Fig. 8 and Fig. 9 show more creative use cases for this system. The first model is reconstructed from a hand-drawn sketch. While the vehicle is less detailed, the network still generalizes well to hand-drawn sketches, even though it was never trained on such input. Fig. 9 shows a vehicle that is created from a blueprint and 3D printed afterwards.

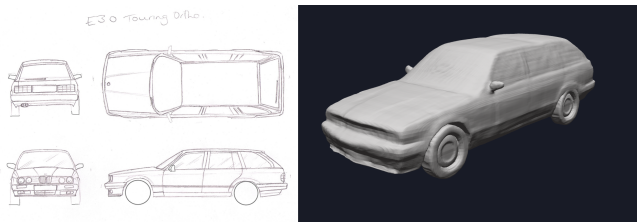

**Figure 8: Reconstruction result from hand drawn sketch, image source: [2]**

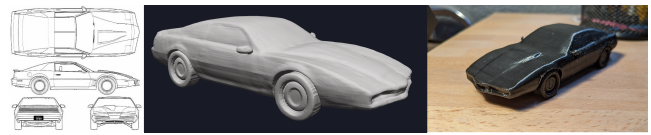

**Figure 9: Reconstructed model after 3D printing**

## 8 ADVANTAGES OVER DEPTH MAP REGRESSION

The implicit function approach has several advantages over depth map regression methods. The first advantage is that there is no limitation on which objects can be reconstructed. Creating a point cloud from depth images always contains areas that are not visible from any camera, which creates empty areas with holes. Applying mesh reconstruction algorithms on the generated point cloud is therefore rather unreliable since there is no guarantee that the result is hole free. The implicit function approach can in theory model any geometry since we can sample signed distance values from any point in space. The reconstructed result is always free of holes since we predict a 3D volume, which is recovered with iso-surface extraction algorithms.

Another problem with depth map regression is the lack of spatial transformation of the input images. Each image is put into its own channel, but the network does not have information on how these images relate to each other in 3D space. This problem is solved with the implicit function approach since each 3D sample point is mapped to the correct two-dimensional point on the image plane for each image.

This allows the network to be trained in three dimensions, while the depth map method is only conditioned to create two-dimensional images. The network does not "know" which post-processing operations are applied afterwards and that the result should be a 3D object. This is the reason why the implicit function approach does not suffer from the view consistency problem.

The final advantage is the possibility to use varying image sizes during the reconstruction. The depth map regression approach can only utilize images with identical sizes since they could not be concatenated in the network otherwise. This causes unnecessary operations and higher VRAM consumption, which is not a problem with the implicit function approach. This is because each image can be encoded separately before we sample features from each image. The different resolutions are irrelevant after the sampling operation, so we can easily concatenate information across all views.

## 9 FAILED EXPERIMENTS

Since the updated PIFuHD [23] variant predicts normal maps from RGB input images as a preprocessing step with a pix2pix [8] network, it is also tested if this step is useful for the task of this paper.

The pix2pix network is trained to predict realistic normals maps from augmented drawings in one experiment. The predictions of the trained network are then used as input for the rest of the pipeline. This slowed down the training process however and did not provide any noticeable improvement. The reasons for this is likely that the predicted normal map is only dependent on a single view and cannot take multi-view information into account. This creates plausible normal maps, but they can be a mismatch to the actual shape. Since the PiFu based network is trained on the ground truth signed distance values, the network cannot "trust" the predictions of the pix2pix network.

The following experiment, therefore, conditioned the PIFu based network on ground truth normal maps, which leads to high-quality reconstructions. The predicted normal maps are then fed into the pre-trained network, which "trusts" the prediction since it is trained on ground truth normal maps. However, since the pix2pix network cannot use multi-view information, the predicted vehicles have inconsistent geometry.

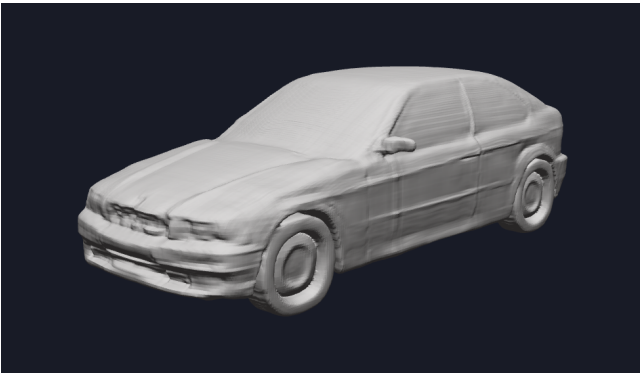

**Figure 10: Reconstruction result after using a pix2pix preprocessing network**

This can be seen in Fig. 10: The hood is pressed in instead of pushed out, which the pix2pix network predicted incorrectly. The result generally has sharper edges and features in some areas but provides less convincing results. An experiment to predict feature maps with the same pix2pix architecture also failed with even less convincing results.

The paper "Implicit Neural Representations with Periodic Activation Functions" [26] suggests using sinusoidal activations (SIREN) inside the multilayer perceptron instead of ReLU activations with a special initialization procedure. The network overfits to the training data however in the deducted experiments. In their experiments, the authors try to fit a signed distance function to a single point cloud, so they do encourage overfitting instead of generalization. It can be seen however that SIRENS have high potential in surface reconstruction if a better training procedure is used.

## 10 CONCLUSION AND FUTURE WORK

It can be seen that pixel-aligned implicit function approaches have high surface quality, high expressiveness in its representation power and create reliable and consistent results. Using pixel-aligned image features allows the network to leverage local information inside the image, compared to methods such as DeepSDF [20], which uses a global latent vector to describe the entire shape.

The result section of the implicit function approach show which results can be expected for real-world inputs. It can be seen that the network generalizes well from synthetic blueprints to their real-world counterparts and creates plausible reconstructions. It is also visible that the network even generalizes to hand-drawn sketches, which are far less precise than technical drawings. This offers new possibilities in the field of rapid prototyping by supporting the creative process of car design via neural networks.

There are plenty of areas in this paper that could be improved upon. The main reason why the reconstructed vehicles do not look perfectly realistic is that they are joined into a single piece. A solution for this could be to train a segmentation classifier, which takes the output from the signed distance prediction regressor as an input and predicts a class label for each point in space. This would require a labelled dataset, however.

The next problem is that the reconstruction tends to be overly smooth in unambiguous areas. This is because the network is trained to regress signed distance values of a ground truth mesh and is therefore punished for any prediction that does not exactly match the original model. Since this is impossible in some areas, the network predicts a smooth transition, because this is the best way to keep the average error low.

This indicates a possible use case for conditional adversarial loss, which has already been explored [13]. The authors use DeepSDF [20] as a base and train an unconditional adversarial network with PointNet [21] as discriminator. It would be interesting to advance this idea to train a conditional adversarial pixel-aligned network with a more recent point cloud network as a discriminator.

Another improvement could be the usage of Dual Contouring [24], a more advanced surface reconstruction algorithm, which should yield sharper edges than Marching Cubes [16].

It would also be possible to apply differentiable rendering, which has been applied on DeepSDF [20] by the paper "Rendering Deep Implicit Signed Distance Function with Differentiable Sphere Tracing" [15]. This makes it possible to condition the network on images of the generated surface, which could improve the fidelity. However, sphere tracing is slow since many iterations have to be processed to compute an image. Using this in a pixel-aligned manner also requires more memory than DeepSDF [20], which uses a global latent vector.